\documentclass[runningheads]{llncs}
\usepackage[T1]{fontenc}
\usepackage{graphicx}
\begin{document}
\title{Cross-Modal Video to Body-joints Augmentation for Rehabilitation Exercise Quality Assessment
}
\titlerunning{Cross-Modal Augmentation for Exercise Assessment}
%

\author{Ali Abedi\inst{1} \and
Mobin Malmirian\inst{2} \and
Shehroz S. Khan\inst{1,3}}

\authorrunning{Abedi et al.}

%

\institute{
KITE Research Institute, University Health Network, Toronto, Canada
\and
Faculty of Applied Science and Engineering, University of Toronto, Canada
\and
Institute of Biomedical Engineering, University of Toronto, Canada\\
\email{\{ali.abedi,shehroz.khan\}@uhn.ca, mobin.malmirian@mail.utoronto.ca}
}

\maketitle              
\begin{abstract}
Exercise-based rehabilitation programs have been shown to enhance quality of life and reduce mortality and rehospitalizations. AI-driven virtual rehabilitation programs enable patients to complete exercises independently at home while AI algorithms can analyze exercise data to provide feedback to patients and report their progress to clinicians. This paper introduces a novel approach to assessing the quality of rehabilitation exercises using RGB video. Sequences of skeletal body joints are extracted from consecutive RGB video frames and analyzed by many-to-one sequential neural networks to evaluate exercise quality. Existing datasets for exercise rehabilitation lack adequate samples for training deep sequential neural networks to generalize effectively. A cross-modal data augmentation approach is proposed to resolve this problem. Visual augmentation techniques are applied to video data, and body joints extracted from the resulting augmented videos are used for training sequential neural networks. Extensive experiments conducted on the KInematic assessment of MOvement and clinical scores for remote monitoring of physical REhabilitation (KIMORE) dataset, demonstrate the superiority of the proposed method over previous baseline approaches. The ablation study highlights a significant enhancement in exercise quality assessment following cross-modal augmentation.

\end{abstract}
\section{Introduction}
The referral of patients who have experienced a stroke, cardiac event, or injury to rehabilitation programs is a widely adopted strategy with the objective of enhancing patients' quality of life and reducing re-hospitalization and mortality rates \cite{dibben2023exercise}. A significant part of these programs focuses on prescribed regular exercises that are designed to facilitate the restoration of mobility and strength among patients \cite{dibben2023exercise}. In recent years, there has been a notable emergence of Artificial Intelligence (AI)-driven virtual rehabilitation as a promising approach for remotely delivering rehabilitation programs to patients within the confines of their own homes \cite{ferreira2023usage}. Studies have shown that virtual rehabilitation can offer health benefits comparable to in-person rehabilitation and that it can overcome several barriers, such as transportation, the cost of therapy, and financial concerns \cite{seron2021effectiveness}. Virtual rehabilitation can involve the utilization of diverse sensors to capture patients' movements, which are subsequently subjected to analysis using AI algorithms during exercise sessions \cite{ferreira2023usage,sangani2020real}. The analysis outcomes derived from these algorithms can be used to provide patients with feedback regarding the quality or completion of their exercises \cite{sangani2020real,fernandez2018virtualgym}. Additionally, clinicians can utilize the analysis results to closely monitor patients' progress and implement appropriate personalized interventions.

Rehabilitation programs typically involve prescribing patients with specific exercises, as well as other components such as education \cite{liao2020review,dibben2023exercise,capecci2019kimore,vakanski2018data,miron2021intellirehabds}. The assessment of exercise performance relies on objective criteria, including adhering to the prescribed exercise sets and repetitions \cite{abedi2023rehabilitation}, maintaining consistency in the execution of exercises, demonstrating proper technique and movement quality, and ensuring the correct posture of various body parts\cite{capecci2019kimore,liao2020review}. Previous research on generic human activity analysis has highlighted the importance of incorporating joint modelling within the parts of the human body \cite{yan2018spatial,liao2020review}. In the context of rehabilitation exercise quality assessment, the incorporation of joint modelling within patients' body parts aligns with the approach taken by clinicians when assessing the quality and technique of exercises \cite{capecci2019kimore}. Body joints exhibit resilience towards variations in illumination and background, making them a reliable data modality for analysis. Body joint information can be acquired through specialized hardware, including depth cameras and wearable sensors \cite{capecci2019kimore,liao2020review}. Alternatively, computer vision techniques can be employed to extract body joint information from RGB videos \cite{pavllo20193d,lugaresi2019mediapipe}. This work specifically concentrates on the latter approach, which eliminates the requirement for costly hardware/equipment and enables the integration of body joint analysis into platforms that solely rely on commonly available RGB cameras. The body joint sequences are extracted from consecutive frames of the videos of exercises. These sequences are then input into a deep sequential neural network for analysis, enabling the inference of exercise quality in the form of a real number.

Obtaining sufficiently large-scale annotated data is often a challenge in many practical scenarios due to the expensive and time-consuming nature of data collection and annotation \cite{athanasiadis2020audio}. This issue is particularly pronounced in healthcare-related contexts and patient data, where acquiring data from patients, such as rehabilitation exercise videos from post-stroke patients at home, poses greater difficulties compared to other applications \cite{capecci2019kimore}. Additionally, annotating patient data incurs higher costs since it requires expert annotators such as physiotherapists and stroke rehabilitation clinicians to score the quality of stroke rehabilitation exercises. To annotate rehabilitation exercises, clinicians rely on video data modality since simply having skeletal body joint data is insufficient for assessing the quality and technique of exercises \cite{capecci2019kimore}. Consequently, while most deep learning models are trained on body joint data modality \cite{liao2020review}, in real-world scenarios, there is typically a higher availability of annotated video data. As a result, the majority of existing exercise rehabilitation datasets lack sufficient samples to be used for training generalizable deep learning models \cite{liao2020review,capecci2019kimore,sardari2023artificial}. The purpose of this work is to overcome this difficulty by applying visual augmentation techniques to annotated videos in order to increase the number of training samples. The main contribution of this work is cross-model data augmentation from videos to body joints; augmenting video data modality, and using the augmented data in the form of body joints data modality to train generalizable deep learning models. Extensive experiments on the only publicly available rehabilitation exercise video dataset, KInematic assessment of MOvement and clinical scores for remote monitoring of physical REhabilitation (KIMORE) \cite{capecci2019kimore}, demonstrated the effectiveness of the cross-modal augmentation and superiority of the proposed method compared to previous methods.

This paper is structured as follows. Section \ref{sec:related_work} introduces related works on rehabilitation exercise quality assessment. In Section \ref{sec:methodology}, the proposed method for rehabilitation exercise quality assessment is presented. Section \ref{sec:experiments} describes experimental settings and results on the proposed methodology. In the end, Section \ref{sec:conclusion} presents our conclusions and directions for future works.

\section{Related Work}
\label{sec:related_work}
In this section, a brief review is conducted on the current state of research pertaining to two key areas: rehabilitation exercise quality assessment and cross-modal data augmentation.

\subsection{Rehabilitation Exercise Quality Assessment}
\label{sec:related_work_exercise}
Liao et al. \cite{liao2020review} classified the methodologies employed for assessing the quality of rehabilitation exercises into three categories: discrete movement scores, rule-based, and template-based approaches. Discrete movement score approaches utilize traditional machine learning classifiers to categorize rehabilitation exercises into discrete groups, such as correct or incorrect \cite{um2018parkinson,vakanski2018data,liao2020review}. These methods, however, are limited in their ability to detect subtle variations in patient performance and provide nuanced assessments of movement quality, such as scores between 0 and 1 \cite{liao2020review}. Rule-based approaches, on the other hand, require clinicians to determine in advance a set of rules for each particular rehabilitation exercise. These rules serve as a benchmark for evaluating an exercise's level of correctness \cite{liao2020review,capecci2019kimore,sardari2023artificial}. The drawback of rule-based approaches lies in the fact that they are exercise-specific and cannot be applied to other exercises \cite{liao2020review,sardari2023artificial}. A template-based approach evaluates exercises against a reference exercise that is correct. Template-based approaches can be categorized as either model-free (direct matching) or model-based \cite{sardari2023artificial,liao2020deep}. Model-free approaches apply a distance function, such as dynamic time warping, between the sequence of movements performed by the patient and the reference sequence of movements of the correct exercise \cite{liao2020review,sardari2023artificial}. Machine-learning and deep-learning approaches are in the category of model-based approaches \cite{liao2020review,sardari2023artificial,liao2020deep,deb2022graph,guo2021exercise}. Some of the model-based approaches extract a single feature vector from the entire video sample/body joint sequence and develop a machine-learning model to analyze the feature vector in a non-sequential manner \cite{guo2021exercise}, but most of them address exercise quality assessment as a spatial-temporal data analysis problem using models such as variants of recurrent neural networks \cite{liao2020deep,liao2020review} and spatial-temporal graph neural networks \cite{liao2020review,sardari2023artificial,deb2022graph}.

\subsection{Cross-Modal Data Augmentation}
\label{sec:related_work_cross}
Data augmentation across modalities has been employed in a variety of applications, including for emotion recognition using image and audio data\cite{wang2021semi}. Wang et al. \cite{wang2021semi} presented an approach for transferring knowledge between annotated facial images and audio domains by learning a joint distribution of samples in different modalities and mapping an image sample to an audio spectrum. Other examples of cross-modal data augmentation include across image and text \cite{wang2022paired}, and video and Lidar data \cite{wang2021pointaugmenting}.

In this paper, we propose the application of cross-modal data augmentation across video and body joint data modalities to increase the number of body joint sequences that can be utilized for training generalizable deep sequential models.

\section{Method}
\label{sec:methodology}
The proposed method takes as input an RGB video sample containing a rehabilitation exercise performed by a patient. The output is a real number that reflects the quality of the rehabilitation exercise. A sequence of body joints is extracted from consecutive frames of the video, through the use of a pre-trained pose estimation model. A sequence of feature vectors is extracted from the sequence of body joints, by extracting hand-crafted features from the body joints in each frame. The features are exercise-specific. For instance, the upper body is given more consideration in the arm-lifting exercise \cite{capecci2019kimore}, where features such as elbow angle and distance between the hands are calculated \cite{guo2021exercise}. To analyze the sequence of feature vectors, a many-to-one Recurrent Neural Network (RNN) is employed. Each feature vector extracted from one frame of the input data is fed to a corresponding timestamp of the RNN. Following the RNN, fully-connected layers are utilized, culminating in a single neuron in the final layer, which generates an output representing exercise quality.

To enhance the number of annotated training samples and improve the generalizability of RNNs, visual augmentation techniques are implemented on the video samples within the training dataset. Care is taken to select video transformation techniques that do not alter the inherent nature of the data or its annotation, specifically the exercise quality depicted in the video. The chosen video transformations encompass horizontal flipping and slight rotations applied uniformly to all video frames. However, temporal data augmentation is not employed on videos to avoid altering the structure of rehabilitation exercises and the way they have been annotated by clinicians \cite{capecci2019kimore}. By utilizing the augmented video data, the corresponding body joints are extracted and employed to train RNNs effectively. The details of the parameters of RNNs are explained in Section \ref{sec:experimental_settings}.

\section{Experiments}
\label{sec:experiments}
This section evaluates the performance of the proposed method on the KIMORE dataset \cite{capecci2019kimore} in comparison to previous methods. The KIMORE dataset \cite{capecci2019kimore} contains RGB and depth videos along with body joint position and orientation data captured by the Kinect camera. The data were collected from 78 participants, including 44 healthy participants and 34 patients with motor dysfunction (stroke, Parkinson’s disease, and low back pain). Each data sample in this dataset is composed of one participant performing multiple repetitions of one of the five exercises: (1) lifting of the arms, (2) lateral tilt of the trunk with the arms in extension, (3) trunk rotation, (4) pelvis rotations on the transverse plane, and (5) squatting. The data samples were annotated by clinical experts in terms of exercise quality as a number in the range of 0 to 50 \cite{capecci2019kimore}. In this dataset, there are 353 samples with an average of 70 samples per exercise, which is quite a few samples for training deep learning models.

\subsection{Experimental Settings}
\label{sec:experimental_settings}
The experiments were conducted using RGB videos available in KIMORE \cite{capecci2019kimore}. MediaPipe \cite{lugaresi2019mediapipe}, a pre-trained deep neural network for pose estimation, was used to extract body-joints sequences from consecutive video frames. Exercise-specific features, designed for five different exercises in KIMORE, were extracted from body joint sequences to construct feature vectors for each video sample. More information about exercise-specific features can be found at \cite{capecci2019kimore,guo2021exercise}. Long Short-Term Memory (LSTM) is used to analyze the sequences of exercise-specific features. The number of neurons in the input layers of the LSTM depends on the dimensionality of features extracted from each exercise \cite{guo2021exercise}. In the LSTM, there are 4 unidirectional layers that contain 16 neurons in hidden layers. Each LSTM layer except the last is followed by a dropout layer with a probability of 0.17. The LSTM is trailed by a $16 \times 1$ linear layer to output the exercise quality score. Since the model solves a regression problem, no activation function was used. The Adam optimization algorithm was used with 300 epochs, batches of size 8, and a learning rate of 0.01 to minimize a mean absolute error loss function \cite{paszke2019pytorch}.

Using samples of the five exercises in KIMORE, five different models were trained and evaluated. The models were trained and evaluated using a five-fold cross-validation procedure. During each iteration of five-fold cross-validation, video augmentation is applied only to the training samples. A total of seven new videos were generated from each video sample by adding horizontal flipping, -1, -2, -3, 1, 2, and 3 degree rotations to the existing video.

\begin{table*}[]
\caption{The Spearman's rank correlation between predictions and ground-truth exercise quality scores in five different exercises in the KIMORE dataset \cite{capecci2019kimore} for different settings of the proposed method compared to previous methods.}
\begin{center}
\includegraphics[scale=.5]{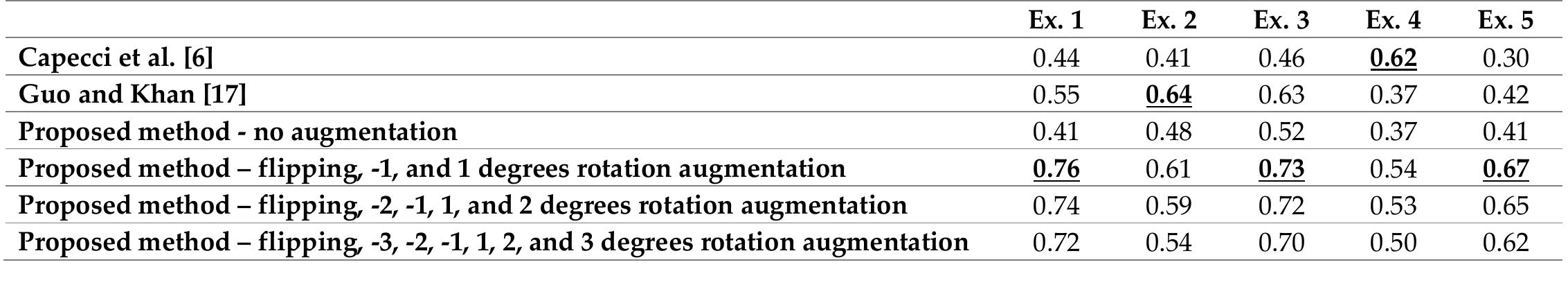}\\
\end{center}
\label{tab:correlation}
\end{table*}

\subsection{Experimental Results}
\label{sec:experimental_results}
Following the literature \cite{liao2020review,liao2020deep,deb2022graph,guo2021exercise}, the Spearman's rank correlation was used as the evaluation metric. The performance of the proposed method in different settings compared to the previous works is shown in Table. \ref{tab:correlation}. It is noteworthy that the samples in the five folds of cross validation in our experiments are not exactly the same samples as those in the previous methods. Compared to no augmentation and with the small number of samples in the individual exercises in KIMORE, cross-modal video-to-body-joint data augmentation significantly improved the performance. For instance, for the first exercise, Ex. 1 in Table. \ref{tab:correlation}, the augmentation resulted in an 86\% improvement in the correlation from 0.41 to 0.76. For the proposed method, the best results were obtained using the first type of data augmentation in which three samples were generated from each video sample by flipping and rotating -1 and 1 degrees. The proposed method significantly outperformed the previous methods on the first, third, and fifth exercises, with 72\%, 59\%, and 123\% improvements in correlation, respectively. While for the second exercise, the performance of the proposed method is superior to \cite{capecci2019kimore} and close to \cite{guo2021exercise}, the performance of the proposed method on the fourth exercise is inferior to \cite{capecci2019kimore}. This is due to the fact that the fourth exercise involved pelvis rotations, i.e., movements along the z axis, which is relatively difficult to capture by the MediaPipe \cite{lugaresi2019mediapipe}.

\begin{figure*}
    \centering
    \includegraphics[scale=.35]{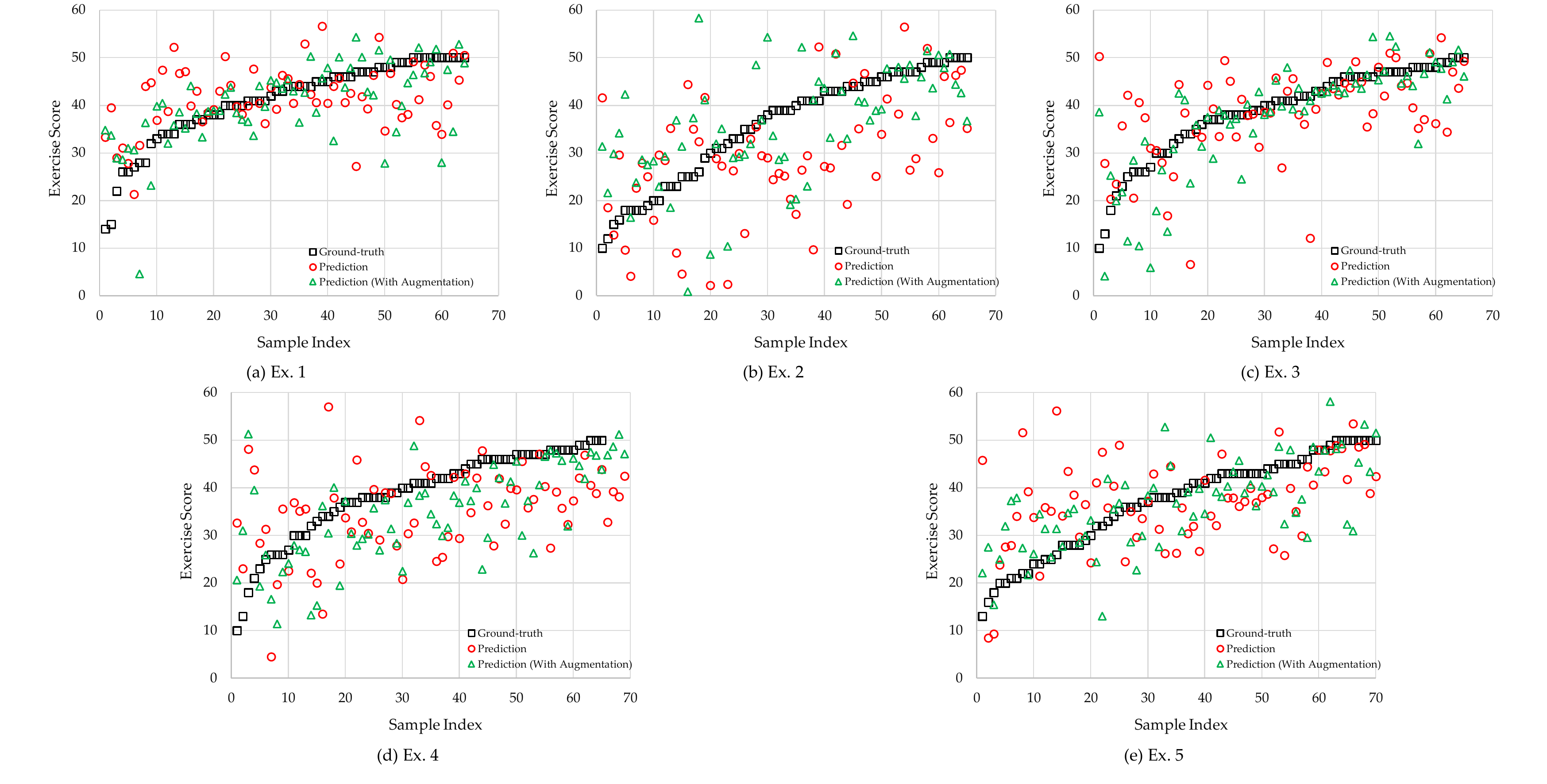}\\
    \caption{The scatter plots of the predictions of the proposed method without and with cross-modal augmentation compared to ground-truth exercise quality scores in (a)-(e) five exercises in the KIMORE dataset \cite{capecci2019kimore}.}
    \label{fig:scatter_plot}
\end{figure*}

Corresponding to the third and fourth rows of Table. \ref{tab:correlation}, Fig. \ref{fig:scatter_plot} illustrates the scatter plots of the proposed method without and with cross-modal augmentation for five exercises in KIMORE. According to Fig. \ref{fig:scatter_plot}, cross-modal augmentation results in predictions that are more closely correlated with ground-truth exercise quality scores.

\section{Conclusion and Future Works}
\label{sec:conclusion}
The purpose of this study was to investigate the effectiveness of cross-modal data augmentation from RGB videos to body joints for developing sequential neural networks for rehabilitation exercise quality assessment. With data augmentation, more generalizable neural networks were trained. This resulted in significant improvement in performance compared to no augmentation and outperformance compared to previous methods. This study represents the first work on video-to-body-joint cross-modal data augmentation \cite{chen2022cross,wang2021semi,wang2022paired,wang2021pointaugmenting}. Future research may involve the inclusion of cross-modal data augmentation within a generative adversarial network setting and using spatial-temporal graph convolutional networks for rehabilitation exercise quality assessment.

%
%
%

\begin{thebibliography}{8}

\bibitem{dibben2023exercise} Dibben, Grace O., et al. "Exercise-based cardiac rehabilitation for coronary heart disease: a meta-analysis." European Heart Journal 44.6 (2023): 452-469.

\bibitem{ferreira2023usage} Ferreira, Ricardo, Rubim Santos, and Andreia Sousa. "Usage of Auxiliary Systems and Artificial Intelligence in Home-Based Rehabilitation: A Review." Exploring the Convergence of Computer and Medical Science Through Cloud Healthcare (2023): 163-196.


\bibitem{seron2021effectiveness} Seron, Pamela, et al. "Effectiveness of telerehabilitation in physical therapy: a rapid overview." Physical therapy 101.6 (2021): pzab053.

\bibitem{sangani2020real} Sangani, Samir, et al. "Real-time avatar-based feedback to enhance the symmetry of spatiotemporal parameters after stroke: instantaneous effects of different avatar views." IEEE Transactions on Neural Systems and Rehabilitation Engineering 28.4 (2020): 878-887.

\bibitem{fernandez2018virtualgym} Fernandez-Cervantes, Victor, et al. "VirtualGym: A kinect-based system for seniors exercising at home." Entertainment Computing 27 (2018): 60-72.

\bibitem{liao2020review} Liao, Yalin, et al. "A review of computational approaches for evaluation of rehabilitation exercises." Computers in biology and medicine 119 (2020): 103687.

\bibitem{capecci2019kimore} Capecci, Marianna, et al. "The KIMORE dataset: KInematic assessment of MOvement and clinical scores for remote monitoring of physical REhabilitation." IEEE Transactions on Neural Systems and Rehabilitation Engineering 27.7 (2019): 1436-1448.

\bibitem{vakanski2018data} Vakanski, Aleksandar, et al. "A data set of human body movements for physical rehabilitation exercises." Data 3.1 (2018): 2.

\bibitem{miron2021intellirehabds} Miron, Alina, et al. "IntelliRehabDS (IRDS)—A dataset of physical rehabilitation movements." Data 6.5 (2021): 46.

\bibitem{yan2018spatial} Yan, Sijie, Yuanjun Xiong, and Dahua Lin. "Spatial temporal graph convolutional networks for skeleton-based action recognition." Proceedings of the AAAI conference on artificial intelligence. Vol. 32. No. 1. 2018.

\bibitem{pavllo20193d} Pavllo, Dario, et al. "3d human pose estimation in video with temporal convolutions and semi-supervised training." Proceedings of the IEEE/CVF conference on computer vision and pattern recognition. 2019.

\bibitem{lugaresi2019mediapipe} Lugaresi, Camillo, et al. "Mediapipe: A framework for building perception pipelines." arXiv preprint arXiv:1906.08172 (2019).

\bibitem{chen2022cross} Chen, Dong, et al. "Cross-modal Data Augmentation for Tasks of Different Modalities." IEEE Transactions on Multimedia (2022).

\bibitem{um2018parkinson} Um, Terry Taewoong, et al. "Parkinson's disease assessment from a wrist-worn wearable sensor in free-living conditions: Deep ensemble learning and visualization." arXiv preprint arXiv:1808.02870 (2018).

\bibitem{sardari2023artificial} Sardari, Sara, et al. "Artificial Intelligence for skeleton-based physical rehabilitation action evaluation: A systematic review." Computers in Biology and Medicine (2023): 106835.

\bibitem{liao2020deep} Liao, Yalin, Aleksandar Vakanski, and Min Xian. "A deep learning framework for assessing physical rehabilitation exercises." IEEE Transactions on Neural Systems and Rehabilitation Engineering 28.2 (2020): 468-477.

\bibitem{deb2022graph} Deb, Swakshar, et al. "Graph convolutional networks for assessment of physical rehabilitation exercises." IEEE Transactions on Neural Systems and Rehabilitation Engineering 30 (2022): 410-419.

\bibitem{guo2021exercise} Guo, Qingyang, and Shehroz Khan. "Exercise-specific feature extraction approach for assessing physical rehabilitation." 4th IJCAI Workshop on AI for Aging, Rehabilitation and Intelligent Assisted Living. IJCAI. 2021.

\bibitem{abedi2023rehabilitation} Abedi, Ali, et al. "Rehabilitation Exercise Repetition Segmentation and Counting using Skeletal Body Joints." arXiv preprint arXiv:2304.09735 (2023).

\bibitem{reby2023graph} Réby, Kévin, et al. "Graph Transformer for Physical Rehabilitation Evaluation." 2023 IEEE 17th International Conference on Automatic Face and Gesture Recognition (FG). IEEE, 2023.

\bibitem{athanasiadis2020audio} Athanasiadis, Christos, Enrique Hortal, and Stylianos Asteriadis. "Audio–visual domain adaptation using conditional semi-supervised generative adversarial networks." Neurocomputing 397 (2020): 331-344.

\bibitem{wang2021semi} Wang, Shaoqiang, et al. "Semi-supervised classification-aware cross-modal deep adversarial data augmentation." Future Generation Computer Systems 125 (2021): 194-205.

\bibitem{wang2022paired} Wang, Hao, et al. "Paired Cross-Modal Data Augmentation for Fine-Grained Image-to-Text Retrieval." Proceedings of the 30th ACM International Conference on Multimedia. 2022.

\bibitem{wang2021pointaugmenting} Wang, Chunwei, et al. "Pointaugmenting: Cross-modal augmentation for 3d object detection." Proceedings of the IEEE/CVF Conference on Computer Vision and Pattern Recognition. 2021.

\bibitem{paszke2019pytorch} Paszke, Adam, et al. "Pytorch: An imperative style, high-performance deep learning library." Advances in neural information processing systems 32 (2019).


\end{thebibliography}
%

\end{document}